\documentclass[a4paper,fleqn]{cas-dc}

\usepackage{graphicx}
\usepackage{soul}
\usepackage{url}
\usepackage{hyperref}
\usepackage{verbatim}
\usepackage{subfigure}
\usepackage{calc}
\usepackage{amstext,amsmath}
\usepackage{stackengine}

\usepackage{array}
\usepackage{amsmath,amssymb} % define this before the line numbering.
\usepackage{xcolor}
\usepackage{paralist}
\usepackage{multirow}

\usepackage{algorithm}  
\usepackage{algpseudocode}  
\renewcommand{\algorithmicrequire}{\textbf{Input:}}

\usepackage{anyfontsize}
\usepackage{eqparbox}
\newdimen{\algindent}
\algnewcommand\LeftComment[2]{%
\hspace{#1\algindent}$\triangleright$ \eqparbox{COMMENT}{#2} \hfill %
}

\newtheorem{remark}{Remark}

\usepackage{xspace}
\usepackage[numbers]{natbib}

\makeatletter
\DeclareRobustCommand\onedot{\futurelet\@let@token\@onedot}
\def\@onedot{\ifx\@let@token.\else.\null\fi\xspace}

\def\eg{\emph{e.g}\onedot} 
\def\ie{\emph{i.e}\onedot} 
 
\def\etc{\emph{etc}\onedot}

\makeatother
\def\tsc#1{\csdef{#1}{\textsc{\lowercase{#1}}\xspace}}
\tsc{WGM}
\tsc{QE}
\begin{document}
\let\WriteBookmarks\relax
\def\floatpagepagefraction{1}
\def\textpagefraction{.001}

% \shorttitle{Data-free Dense Depth Distillation}    
 \shorttitle{Dense Depth Distillation with Out-of-Distribution Simulated Images}
\shortauthors{Junjie Hu et~al.}  

% \title [mode = title]{Data-free Dense Depth Distillation}  
\title [mode = title]{Dense Depth Distillation with Out-of-Distribution Simulated Images}
% {Data-Free Knowledge Distillation for Monocular Depth Estimation with Out-of-Distribution Simulated Images}  



\author[1]{Junjie Hu}[]
\ead{hujunjie@cuhk.edu.cn}
\address[1]{Shenzhen Institute of Artificial Intelligence and Robotics for Society, Shenzhen, China}

\author[2]{Chenyou Fan}[]
\ead{fanchenyou@gmail.com}
\address[2]{ South China Normal University, China}
 
\author[3]{Mete Ozay}[]
\ead{meteozay@gmail.com}
\address[3]{ METU, Turkey}
 
\author[4]{Hualie Jiang}[]
\ead{hl.jiang@siat.ac.cn}
\address[4]{ The School of Science and Engineering, the Chinese University of Hong Kong, Shenzhen, China}

\author[4,1]{Tin Lun Lam}[]
\cormark[1]
\ead{tllam@cuhk.edu.cn}

\cortext[1]{Corresponding author}



\begin{abstract}
We study data-free knowledge distillation (KD) for monocular depth estimation (MDE), which learns a lightweight model for real-world depth perception tasks by compressing it from a trained teacher model while lacking training data in the target domain. Owing to the essential difference between image classification and dense regression, previous methods of data-free KD are not applicable to MDE. To strengthen its applicability in real-world tasks, in this paper, we propose to apply KD with out-of-distribution simulated images. The major challenges to be resolved are i) lacking prior information about scene configurations of real-world training data and ii) domain shift between simulated and real-world images. To cope with these difficulties, we propose a tailored framework for depth distillation. The framework generates new training samples for embracing a multitude of possible object arrangements in the target domain and utilizes a transformation network to efficiently adapt them to the feature statistics preserved in the teacher model. Through extensive experiments on various depth estimation models and two different datasets, we show that our method outperforms the baseline KD by a good margin and even achieves slightly better performance with as few as $1/6$ of training images, demonstrating a clear superiority. 
\end{abstract}


\maketitle

\section{Introduction}
As a cost-effective alternative solution to depth sensors,
monocular depth estimation (MDE) predicts scene depth from only RGB images and has wide applications in various tasks, such as scene understanding \cite {max-S-and-D}, autonomous driving \citep{song2021self}, 3D reconstruction \citep{XiyueGuo2021}, and augmented reality \citep{Du2020DepthLab}. In recent years, the accuracy of MDE methods has been significantly boosted and dominated by deep learning based approaches \citep{fu2018deep,Hu2019RevisitingSI,jiang2021plnet}, where the advances are attributed to modeling and estimating depth by complex nonlinear functions using large-scale deep neural networks.

On the other hand, many practical applications, \eg, robot navigation, demand a lightweight model due to the hardware limitations and requirement for computationally efficient inference. 
% {\color{blue}
In these cases, we can either perform model compression on a well-trained large network \citep{Wofk2019FastDepthFM} or apply supervised learning to directly train a compact network \citep{Mancini2016FastRM}. These solutions assume that the original training data of the target domain is known and can be freely accessed. However, since data privacy and security are invariably a severe concern in the real world, the training data is routinely unknown in practice, especially for industrial applications. A potential solution under this practical constraint is to distill preserved knowledge from a well-trained and publicly available model without accessing the original training data. The task is called data-free knowledge distillation (KD) \citep{Lopes2017DataFreeKD} and has been shown to be effective for image classification.

\begin{figure}[t!]
    \centering
    \includegraphics[width = 1\linewidth]{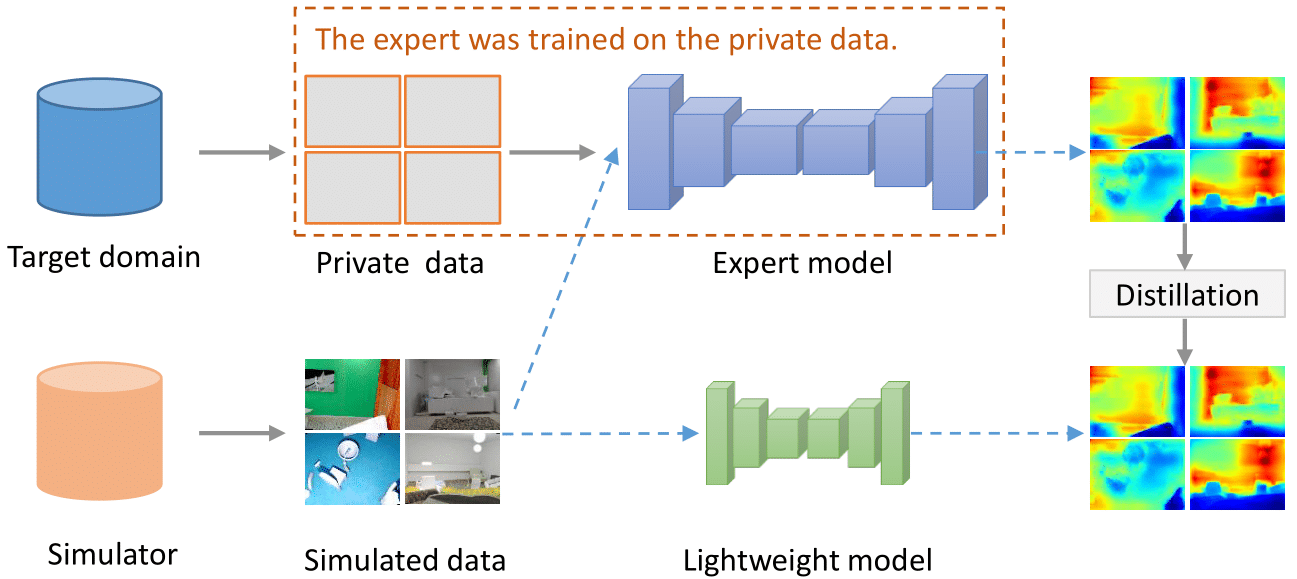}
    \vspace{-3mm}
    \caption{A visualization of the problem of data-free depth distillation. We propose to use simulated images as an alternative solution to the challenges of applying knowledge distillation for monocular depth estimation when original training data is not available.}
    \label{fig-intro}
\end{figure}

Most existing methods of data-free KD proposed to synthesize training images from random noise \citep{Fang2019DataFreeAD,yin2020dreaming}.  Specifically, assuming that $y$ is a target object attribute, it is an element that inherently exists in the last layer of a classifier and is easily pre-specified, such that we can
enforce a classifier to produce the desired output by gradually optimizing its input data. We refer to this property as the inherent constraint of classification. 
Unfortunately, in the case of MDE, the output is a high-resolution two-dimensional map with interrelated objects, not a score for a category; the inherent constraint does not hold for MDE, making most existing data-free approaches incompatible.  More formal analyses are given in Sec.~\ref{premilinary_1}.

\begin{figure*}[t]
    \centering
    \includegraphics[width = 0.92\linewidth]{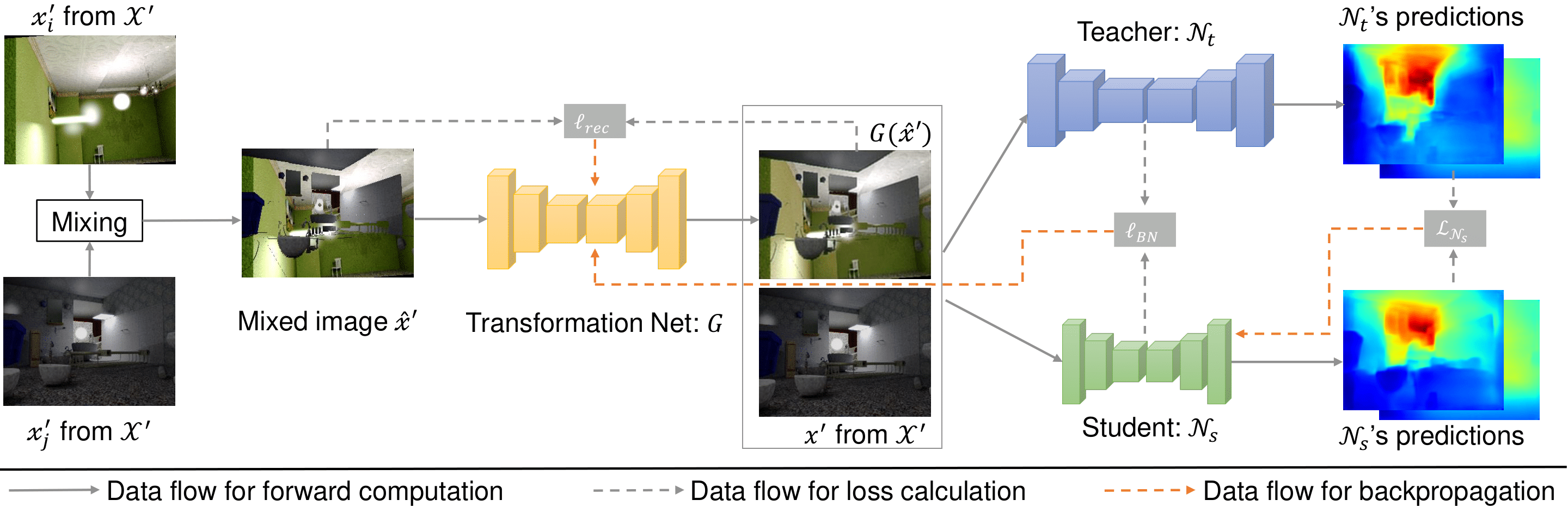}
    % \vspace{-2mm}
    \caption{ A flowchart of the proposed approach for distilling a trained model in the real world with simulated images. 
    We firstly mix two images $x'_i$ and $x'_j$ sampled from the simulated dataset $\mathcal{X'}$ to generate a new sample $\hat{x}'$, and use a transformation network to fit $\hat{x}'$ to the feature distribution provided by the trained teacher model. The distillation is applied from the teacher $\mathcal{N}_t$ to the target student $\mathcal{N}_s$ with the new input $G(\hat{x}')$ and the original simulated data $x'$ where the latter ensures a lower bound performance. The transformation and the student network are trained jointly.}
    \label{fig-dragram}
\end{figure*}
Given the above challenges, in this paper, we propose to leverage out-of-distribution (OOD) images as an alternative solution for applying KD. 
For the MDE task, intuitively, we consider three critical elements for choosing the alternative set: i) the similarity in scene structures between the original scenarios and the OOD set, ii) the number of training images of the OOD set, and iii) domain gap between the original domain and the OOD domain.
We analyze the effect of the above three factors on the accuracy of KD through quantitative experiments. Unsurprisingly, high scene similarity, sufficient data, and a small domain gap contribute to better accuracy. Another valuable observation is that the teacher still estimates meaningful depth maps correctly representing relative depths among objects, even from simulated images. 
It reveals that DNNs may utilize some geometric cues \citep{hu2019visualization,hu2019analysis,Dijk2019HowDN}, or can learn some domain-invariant features \citep{Chen2021S2RDepthNetLA} for inferring depths rather than the straightforward fitting. Therefore, it is still possible to perform KD even though the predicted depth maps are completely wrong in scales.

The effective yet impractical solution is to collect a dataset similar to the original training data. 
In reality, data collection is always costly and time-consuming. Besides, due to the lack of prior information about scene structures of the original training data, we have no sufficient clues to guide this data collection process.
For these reasons, we prefer using synthetic images collected from simulators, as visualized in Figure~\ref{fig-intro}.
In this way, we can handily collect enough images to ensure the multitude of data for distillation.
However, we need to overcome the significant domain gap between simulated and real-world data. 

In general, the difficulties of depth distillation utilizing simulated data are two-fold. The first is the unknown scene/object configurations in the target domain. The second is the unavoidable domain discrepancy between the simulated and original (real-world) training sets. 
Our distillation framework is composed of two sub-branches. The first branch applies the plain KD using initial simulated images to ensure a lower-bound performance. The second branch expands our training set by generating additional training samples to tackle the above challenges. Specifically, we first generate new images that aim to encompass a diverse array of scene configurations in the target domain by applying random object-wise mixing between two simulated images. 
Then, we propose to regularize the mixed images to fit the target domain by tackling an efficient image-to-feature adaption problem with a transformation network. 
Figure~\ref{fig-dragram} shows the diagram of the proposed distillation framework where we learn the transformation network $G$ and the target student network $\mathcal{N}_s$ simultaneously.

To the best of our knowledge, we are the first to distill knowledge for MDE in data-free scenarios. We extensively evaluate the proposed method on different depth estimation models and two indoor datasets, including NYU-v2 and ScanNet. In all datasets, our approach demonstrates the best performance. It outperforms the baseline KD by a good margin, \eg, gaining 0.05 and 0.08 average improvements in RMSE (meters) on NYU-v2 and ScanNet, respectively, and shows slightly better performance with as few as $1/6$ of the image dataset.

In summary, our contributions include:
\begin{enumerate}

    \item The first attempt performing data-free KD for monocular depth estimation using OOD simulated images with quantitative studies to ascertain the essential requirements for selecting the OOD data.

    \item A specialized framework for efficient depth distillation by learning image-to-feature adaption. We propose to use a transformation network for fast distillation.

    \item  We validate the proposed method for different depth estimation models on multiple datasets. As a result, our method outperforms the baseline methods by a good margin.
    
\end{enumerate}
    
The rest of the paper is organized as follows. In Section.~\ref{related_work}, we introduce the related works. In Section.~\ref{premilinary_1}, we give formal analyses regarding the difficulties of applying data-free KD for MDE. Section.~\ref{method} presents our method in detail. Section~\ref{experiments} shows detailed experimental settings and results to verify the effectiveness of our method. Section.~\ref{conclusion} concludes the paper.

\section{Related Work}
\label{related_work}
\subsection{Monocular Depth Estimation}
Monocular depth estimation (MDE) aims to predict scene depths from only a single image. 
Deep learning-based approaches have dominated recent progress \citep{laina2016deeper,ma2017sparse,fu2018deep,huynh2020guiding,jiang2021plnet,yin2019enforcing,yin2021virtual,Zhao2021MaskedGF,Hu2022DeepDC} in which the advanced performances are attributed to modeling and estimating depth using large and complex networks with data-driven supervised/unsupervised learning. 

On the other hand, deploying MDE algorithms into real-world applications often faces practical challenges, such as limited hardware resources and inefficient computation. Therefore, an emerging requirement of MDE is to develop lightweight models to meet the above demands. This problem has been specifically considered in previous studies \citep{Mancini2016FastRM,Nekrasov2019RealTimeJS,Wofk2019FastDepthFM,hu2021boosting} where several different lightweight networks have been designed.

However, lightweight networks inevitably degrade their MDE performance due to the trade-off between model complexity and accuracy. Hence, it remains an open question: how can the model complexity be reduced while maintaining high accuracy? One potential solution to this problem is KD, which transfers the knowledge from a cumbersome teacher network to a compact student network with decent accuracy improvement. However, KD requires the original training dataset for implementation. Currently, there are no existing solutions in data-free scenarios for MDE.

\subsection{Knowledge Distillation}
Knowledge distillation (KD) \citep{Hinton2015DistillingTK} was initially introduced in image classification, where either the soft label or the one-hot label predicted by the teacher is used to supervise student training. 
Existing methods can be generally categorized into two groups depending on whether they can access the original training set: 1) standard data-aware KD and 2) data-free KD.

The effectiveness of data-aware KD has been demonstrated for various vision tasks, such as image classification \citep{Hinton2015DistillingTK}, semantic segmentation \citep{hu2022progressive,Liu2020StructuredKD}, object detection \citep{chen2017learning}, and depth estimation \citep{Pilzer2019RefineAD,Wang2021KnowledgeDF}, \etc. In addition to the conventional setup, researchers have proposed to improve KD via distilling intermediate features \citep{Huang2017LikeWY,Liu2020StructuredKD}, distilling from multiple teachers \citep{Tarvainen2017MeanTA,Liu2019KnowledgeFI}, employing an additional assistant network \citep{Mirzadeh2020ImprovedKD}, and adversarial distillation \citep{Chung2020FeaturemaplevelOA,Shen2019MEALME}.

For data-free KD, researchers resorted to synthesizing training sets from random noise \citep{Lopes2017DataFreeKD,yin2020dreaming,Fang2019DataFreeAD,Fang2021UpT1,Yoo2019KnowledgeEW} or employed other large-scale data from different domains \citep{Chen2021LearningSN,Xu2019PositiveUnlabeledCO,Fang2021MosaickingTD,nayak2021effectiveness}.
However, existing methods are only effective for classification tasks and cannot be applied to MDE. 
% We will elaborate on these observations in Sec.~\ref{premilinary}. 
In this paper, we propose the first method of data-free distillation for MDE.
Our method leverages data from simulated environments to distill a model trained on a real-world dataset.

We would like to emphasize the distinctions between knowledge distillation (KD) and domain adaptation (DA), as there may be some potential misunderstandings. 
Primarily, Domain Adaptation (DA) entails the transfer of a model from its original domain (source domain) to a novel and distinct domain (target domain). In contrast, Knowledge Distillation (KD) pertains to a single domain context, concentrating on upholding the model's accuracy within the original domain. The scenario shifts when training images from the original domain remain undisclosed and inaccessible, leading KD and DA to manifest as specific instances of data-free KD and source data-free DA \citep{kim2021domain,yang2021exploiting}, correspondingly. Nevertheless, it remains noteworthy that DA necessitates access to data from the target domain.

\section{Preliminary Analyses}
\label{premilinary_1}
Our target is to perform data-free KD for MDE tasks, facilitating the development of a lightweight MDE model. We commence by delineating the challenges inherent in this task through meticulous analysis and elucidate why existing methods are unsuitable for MDE. Subsequently, we endeavor to leverage OOD data, supported by quantitative analyses, to discern the critical requisites for effectively incorporating OOD data into the KD framework.

\subsection{Difficulty of Data-Free Depth Distillation}
\label{premilinary_difficulty}
Suppose that $\mathcal{N}_t$ is a  model trained using data from the target domain $\mathcal{D}=\{\mathcal{X}, \mathcal{Y}\} $ where
$\mathcal{X}$ and $\mathcal{Y}$ denote input data (\ie, image) and label space, respectively.
For any $x \in \mathcal{X}$, its corresponding label is estimated by $y=\mathcal{N}_t(x)$.

KD aims at learning a smaller network $\mathcal{N}_s$ with the supervision from $\mathcal{N}_t$. Usually, $\mathcal{N}_t$ is called the teacher network and $\mathcal{N}_s$ is called the student network, respectively.
% such that
Then, the learning is formulated as:
\begin{equation}
    \underset{\mathcal{N}_s} \min    \sum_{x \in \mathcal{X}, y \in \mathcal{Y}} \lambda \mathcal{H} \left( \mathcal{N}_t(x),\mathcal{N}_s(x) \right) + (1 - \lambda) \mathcal{H} \left( y,\mathcal{N}_s(x) \right),
    \label{eq_kd}
\end{equation}
where $\mathcal{H}$ is a loss function, $\lambda > 0$ is a weighting coefficient and usually is a relatively large number, \eg, 0.9, for giving more weights to the teacher predictions than ground truths.
In practice, the second term of Eq. \eqref{eq_kd} is sometimes discarded. In these cases, Eq. \eqref{eq_kd} is simplified using $\lambda=1$ by 
\begin{equation}
    \underset{\mathcal{N}_s} \min   \sum_{x \in \mathcal{X}} \mathcal{H} \left( \mathcal{N}_t(x),\mathcal{N}_s(x) \right).
    \label{eq_kd2}
\end{equation}

As shown above, the standard KD requires accessing the original training data sampled from $\mathcal{X}$. Contrarily,
data-free KD attempts to train the student model without being aware of $\mathcal{X}$. It is formulated by
\begin{equation}
    \underset{\mathcal{N}_s} \min   \sum_{x' \in \mathcal{X'}} \mathcal{H} \left( \mathcal{N}_t(x'),\mathcal{N}_s(x') \right),
    \label{eq_free_kd}
\end{equation}
where $\mathcal{X'}$ is a proxy to $\mathcal{X}$ and can be either i) a set of images synthesized from $\mathcal{N}_t$, or ii) other alternative OOD datasets.
Then, Eq.~\eqref{eq_free_kd} can be solved by searching for the optimal $\mathcal{X'}$.

For image classification, the success is attributed to an inherent constraint for identifying $\mathcal{X'}$.
 As $y$ denotes an object category, it corresponds to an index of the SoftMax outputs from the last fully convolutional layer of the model and thus provides prior information about the desired model output.
Then,  $\mathcal{X'}$ is constructed by 
\begin{equation}
  \mathop{\arg\min}\limits_{x'} \sum_{x' \in \mathcal{X'}}    \mathcal{H} ( \mathcal{N}_t(x'),y) + \mathcal{R}(x'),
    \label{eq_free_kd2}
\end{equation}
where $\mathcal{R}$ denotes regularization terms.

\begin{remark}\label{proof1}
The first term of Eq.~\eqref{eq_free_kd2} is an inherently strong constraint of image classification that enforces the output consistency such that two posterior probabilities satisfy
 $\mathcal{P}(y|x) \approx \mathcal{P}(y|x')$.
\end{remark}

\begin{table*}[!th]
\renewcommand\arraystretch{1.2}
\begin{center}
\caption{
Accuracy of the student model employed on the NYU-v2 test set. The student model is trained via knowledge distillation with different OOD data. Except for (g), all datasets have approximately 50K images.}
\label{res_a}
\footnotesize
\begin{tabular}
% {lccccc}
{p{0.02\textwidth}<{\centering}|p{0.15\textwidth}<{\centering}|p{0.1\textwidth}<{\centering}p{0.1\textwidth}<{\centering}p{0.05\textwidth}<{\centering}|p{0.05\textwidth}<{\centering}}
\hline
 & Dataset  &Scene & Domain & Data &$\delta_1$ \\   \hline
(a)& NYU-v2 \citep{NYUv2} &indoor scene &real world  &50K &\textbf{0.808} \\ 
(b)& ImageNet \citep{Deng2009ImageNetAL}  & single object & real wrold&50K &0.685 \\
(c)& Random noises & - & - &50K &0.194 \\
(d)& ScanNet \citep{dai2017scannet} &indoor scene  &real world &50K &0.787  \\
(e)& KITTI \citep{Uhrig2017THREEDV} &outdoor scene &real world  &50K &0.705  \\
(f)& SceneNet \citep{mccormac2016scenenet}&indoor scene &simulation  &50K  &0.712\\ 
(g)& SceneNet \citep{mccormac2016scenenet} &indoor scene &simulation  &300K  &0.742\\ 
\hline
\end{tabular}
\end{center}
\end{table*}

\begin{figure*}[t]
    \centering
    \includegraphics[width = 0.92\linewidth]{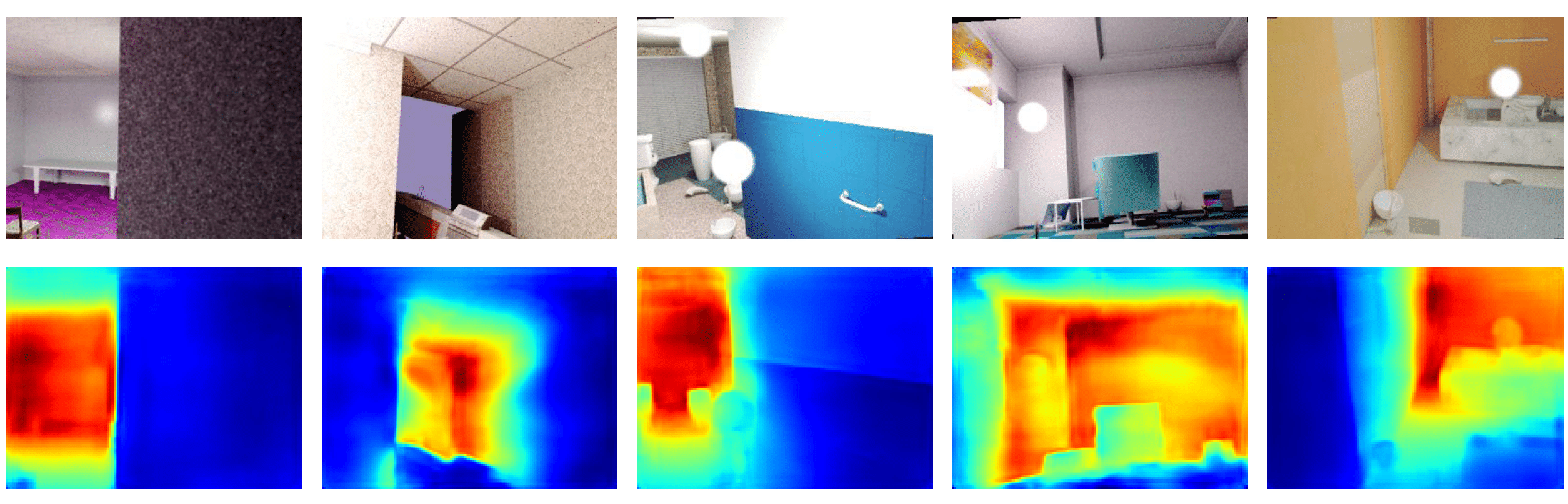}
    \caption{Visualization of the simulated images and the depth maps estimated by the teacher model.}
 \label{fig-res-teacher}
\end{figure*}

We can specify any category corresponding to an actual label of $\mathcal{Y}$ and generate sufficient images from random noises. Besides, in some works, this inherent constraint is used to transform the OOD data to the target distribution \citep{Fang2021MosaickingTD} or identify the most relevant data with low entropy from a large-scale dataset to the distribution of the target domain for efficient KD \citep{Chen2021LearningSN}.

Based on the above discussions, existing approaches of data-free KD on image classification cannot be deployed into our depth estimation task, as predicted depth maps are 2D high-resolution maps with interrelated objects but not score values, invalidating the strong constraint used in Eq.~\eqref{eq_free_kd2}.
Since directly synthesizing data from noise is intractable, we have to seek alternative data to perform KD.

\subsection{Depth Distillation with OOD Data}

According to Remark~\ref{proof1},
data-free KD for MDE can be intractable. 
A plausible way is to use some OOD data if we can decipher the essential requirements for $\mathcal{X'}$. 
Here, we consider that three factors are essential for selecting $\mathcal{X'}$: i) scene structure similarity to $\mathcal{X}$, ii) data-scale for performing KD, and iii) domain gap between $\mathcal{X'}$ and $\mathcal{X}$.

We conducted preliminary experiments to analyze how a depth estimation model reacts to different types of OOD data. We employ a model trained on the NYU-v2 \citep{NYUv2} dataset as the teacher model and apply KD using different OOD datasets with the same number of randomly sampled images.

We observed that the depth range of the depth maps predicted from different types of OOD data still fits into the target domain.
However, this constraint is insufficient to ensure KD, as shown in Table~\ref{res_a}, where random Gaussian noises led to the lowest performance even though they yield a similar depth range to other types of OOD data.

Then, we analyze the effect of the above three factors by comparing scene similarity, data scale, and data domain. Except for $(g)$, all datasets have 50,000 (50K) images. By closely comparing (d) and (e), (d) and (f), and (f) and (g), with (a), not surprisingly, we observe that high scene similarity, small domain gaps in images, and large-scale training datasets are beneficial for the performance boost.

However, it is challenging to satisfy all these three conditions simultaneously. Considering the difficulties of data collection in real-world applications, we propose to apply data-free KD for MDE with simulated images. 
Figure~\ref{fig-res-teacher} shows depth maps estimated from simulated images. It is observed that the inferred depth maps are perceptually correct despite being wrong in absolute depth scales, providing a strong prior of data-free distillation for MDE with simulated data.  We posit that this property is pivotal to the success of depth distillation utilizing simulated data.
% Since there is a clear domain gap between the simulated data and the original real-world data, we need to mitigate the domain gap to promote KD as in Section~\ref{LTFD}.

\section{Framework for Dense Depth Distillation with Simulated Data}
\label{method}

 Our distillation framework consists of three models: the fixed teacher, the target student, and a data transformation model, where the latter two models are jointly trained. We present the learning objectives of the transformation model and the student model as follows.

\subsection{Training the Transformation Model}
\label{LTFD}

We have shown that the trained model would 
estimate reasonable depth maps with correct relative distances among objects from the OOD simulated data. 
However, due to the domain gap, there is a significant discrepancy between $\mathcal{X}$ and $\mathcal{X'}$.
Thus, we wish to mitigate this domain gap and accordingly improve KD.

Since we have no clues about the original training data, such as identities or representations of objects, we naturally consider applying data augmentation to maximally encompass the configurations of scenarios in the target domain. Inspired by ClassMix \citep{Olsson2021ClassMixSD}, we randomly change half of objects between two simulated images to obtain a new image with the help of semantic maps collected from the simulator.
More formally, for two images $x'_i$ and $x'_j$ where $x'_i \in \mathcal{X}'$, $x'_j \in \mathcal{X}'$, we generate a new mixed image $\hat{x}'$ by
\begin{equation}
    \hat{x}' = m \odot x'_i + (1- m) \odot x'_j,
    \label{eq_mix}
\end{equation}
where $m$ is a binary mask obtained from the semantic map of $x'_i$, and
randomly selects half of the classes observed in $x'_i$.

We leverage the running average statistics captured inside neural networks as DeepInversion \citep{yin2020dreaming} to regularize $\hat{x}'$. Specifically, assuming that feature statistics follow the Gaussian distribution and can be defined by mean $\mu$ and variance $\sigma^2$,
then, $\hat{x}'$ is optimized through the following loss
\begin{equation}
   \mathcal{\ell}_{BN} = \sum_{l \in [L]}\|u_l(\hat{x}') - \overline{u}_l \|_2 + \sum_{l \in [L]}\|\sigma_l^2(\hat{x}') - \overline{\sigma}^2_l \|_2,
    \label{eq_bn}
\end{equation}
% The main idea is to 
where $u_l(\hat{x}')$ and $\sigma_l^2(\hat{x}')$ are the batch-wise mean and variance of feature maps of the $l$-th convolutional layer of $\mathcal{N}_t$, respectively. $\overline{u}_l$ and $\overline{\sigma}^2_l$ are the running mean and variance of the $l$-th BN layer of $\mathcal{N}_t$, respectively.
Eq.\eqref{eq_bn} allows regularizing $\hat{x}'$ to fit the feature distribution provided by the teacher model. However, this optimization requires thousands of iterations\footnote{3000 iterations are used for DeepInversion.} for a single batch and is highly time consuming. To tackle this problem, we propose to turn the optimization process for estimating $\hat{x}'$ into a representation learning problem by
training an additional model $G$ for data transformation. Then, Eq.\eqref{eq_bn} can be rewritten by
\begin{equation}
   \mathcal{\ell}_{BN} = \sum_{l \in [L]}\|u_l(G(\hat{x}')) - \overline{u}_l \|_2 + \sum_{l \in [L]}\|\sigma_l^2(G(\hat{x}')) - \overline{\sigma}^2_l \|_2.
    \label{eq_bn2}
\end{equation}

\begin{algorithm}[t]  
  \caption{Depth Distillation Algorithm.}  
  \label{alg_leaning_G}  
  \begin{algorithmic}[1]  
    \Require  
      $\mathcal{X'}$: OOD images collected from a simulator;
      $\mathcal{N}_t$: The teacher model trained on a target domain;
      $\alpha$, $\beta$: Weighting coefficients used for defining loss in training $G$; $T$: Number of iterations; 
     \renewcommand{\algorithmicrequire}{\textbf{Hyper-parameters:}}
    \Ensure  $\mathcal{N}_s$: The student model; $G$: The transformation model.
    \State Freeze $\mathcal{N}_t$;
        \State Initialize $\mathcal{N}_s$ and $G$;
        \For{$j$ = 1 to $T$}             
            \State Set gradients of $\mathcal{N}_s$ and $G$ to 0;
            \State Select a batch $x'$ from  $\mathcal{X'}$;
            \State Let $x'_i=x'$ and  $x'_j= \text{random\_shuffle}(x')$;
            \State Generate mixed images $\hat{x}'$ by Eq.~\eqref{eq_mix}; 
            \Statex \LeftComment{1} {\color{gray} \% Updating the student model \%}
            \State Calculate $\mathcal{N}_t(x')$, $\mathcal{N}_s(x')$, $\mathcal{N}_t(G(\hat{x}'))$, $\mathcal{N}_s(G(\hat{x}'))$;
            \State Calculate the depth loss by Eq.~\eqref{eq_learn_S};  
             \State Update $\mathcal{N}_s$; 
            \Statex \LeftComment{1} {\color{gray} \% Updating the transformation network \%}
            \State Calculate the loss for training $G$ by Eq.~\eqref{eq_learn_G};  
            \State Update $G$; 
         \EndFor           
  \end{algorithmic}  
\end{algorithm}

It is essential to ensure the fidelity of the original scenes to keep the geometry structure. Thus, we adopt an image reconstruction loss. 
Finally, the loss function for training the transformation network is defined by
\begin{equation}
    \mathcal{L}_{G} =  \sum_{\hat{x}' \in \mathcal{X'}} \left ( \alpha \mathcal{\ell}_{BN} + \beta \mathcal{\ell}_{rec} \right),
    \label{eq_learn_G}
\end{equation}
where $\mathcal{\ell}_{rec} = \|\hat{x}' - G(\hat{x}') \|_1$ is the reconstruction error that penalizes the $\ell_1$ norm of image difference, and $\alpha$ and $\beta$ are weighting coefficients.

\subsection{Training the Student Model}
We formally describe the distillation framework to enable data-free student training.  We amalgamate the prediction loss from both the teacher and student models using the initial simulated images $x'$ without amalgamation and the transformed images $G(\hat{x}')$ derived from the mixed images $\hat{x}'$.
The former adopts plain distillation to ensure a lower bound performance, and the latter contributes to further performance improvement.

The optimization objective of depth distillation from the teacher model to the student model is defined by

\begin{equation}
\begin{gathered}
     \mathcal{L}_{\mathcal{N}_s} =  \sum_{x' \in \mathcal{X'}} \mathcal{H} \left( \mathcal{N}_t(x'),\mathcal{N}_s(x') \right) + \\ \sum_{\hat{x}' \in \mathcal{\hat{X}'}} \mathcal{H} \left( \mathcal{N}_t(G(\hat{x}')),\mathcal{N}_s(G(\hat{x}'))\right),
\end{gathered}
\label{eq_learn_S}
\end{equation}
where $\mathcal{H}$ is a function used for measuring the depth errors. We employ the function proposed in \citep{Hu2019RevisitingSI} that penalizes losses of depth, gradient, and normal.

 To facilitate efficient learning, the transformation model and the student model are trained jointly by solving the following optimization problem
\begin{equation}
   \underset{G,\mathcal{N}_s} \min (\mathcal{L}_{G} +  \mathcal{L}_{\mathcal{N}_s})
    \label{eq_learn_G_S}.
\end{equation}
The details of our method are given in Algorithm~\ref{alg_leaning_G} where random\_shuffle denotes the operation of randomizing images.

\section{Experimental Analyses}
\label{experiments}

\subsection{Experimental Settings}

\begin{table}[!tbh]
\caption{Details of the RGBD datasets used in the experiments.}
\begin{center}
\label{datasets}
\footnotesize
\begin{tabular}
{p{0.12\textwidth}<{\centering}p{0.1\textwidth}<{\centering}p{0.085\textwidth}<{\centering}p{0.085\textwidth}<{\centering}}
\hline
&\multirow{3}{*}{Dataset}  & Training & Test \\ 
& & scenarios / images & scenarios / images \\ \hline
\multirow{4}{*}{Target domain}  & \multirow{2}{*}{NYU-v2} & \makecell[c]{249 / \\  50688} &  \makecell[c]{215 / \\   614}\\ 
&  \multirow{2}{*}{ScanNet} &1513 / 50473 &100 / 17607\\ \hline
\multirow{4}{*}{Simulated data} &  \multirow{2}{*}{SceneNet $\mathcal{X}'_1$} &\makecell[c]{1000 / \\   50K}&  -\\
&  \multirow{2}{*}{SceneNet $\mathcal{X}'_2$} &1000 / 300K & -\\
% & SceneNet $\mathcal{X}'_3$ &2000 / 100K & -\\
\hline
\end{tabular}
\end{center}
\end{table}

\subsubsection{Implementation Details}
Our learning framework includes three models: (1) a teacher model $\mathcal{N}_t$ trained on a given target domain and is fixed while training a student model; (2) a student model $\mathcal{N}_s$, which we aim to train; and (3) a transformation network $G$ which will also be optimized during training. 
We train $\mathcal{N}_s$ and $G$ for 20 epochs using the Adam optimizer \citep{kingma2014adam}.  The learning rate begins at 0.0001 and is halved every five epochs. The hyper-parameters $\alpha$ and $\beta$ controlling the data transformation are set to 0.001 for all experiments throughout the paper.
 We trained models with a batch size of 8 in 
all the experiments and developed the code-base using PyTorch \citep{NEURIPS2019_9015}. 

\begin{table*}[t]
\caption{Quantitative results on the NYU-v2 dataset. We use two popular measures including the root mean squared error (RMSE) and thresholding accuracy $\delta_1$.}
\renewcommand\arraystretch{1.3}
\begin{center}
\label{res_nyu}
\footnotesize
\begin{tabular}
{cc|cc|cc|cc|cc|cc}
\hline
\multicolumn{2}{r|}{Model} & \multicolumn{2}{c|}{ LDEN \citep{hu2021boosting} }& \multicolumn{2}{c|}{ LDEN \citep{hu2021boosting}} &\multicolumn{2}{c|}{ EDN \citep{laina2016deeper}} &\multicolumn{2}{c|}{ FFN \citep{Hu2019RevisitingSI}} &\multicolumn{2}{c}{ SARPN \citep{Chen2019structure-aware}} \\ 

\hline
\multicolumn{2}{r|}{ \makecell[r]{Teacher (Backbone)\\   $\rightarrow$  Student (Backbone)} }& \multicolumn{2}{c|}{  \makecell[c]{ResNet-34 \\  $\rightarrow$ ResNet-34}}  & \multicolumn{2}{c|}{  \makecell[c]{ResNet-34\\  $\rightarrow$ MobileNet-v2}} & \multicolumn{2}{c|}{  \makecell[c]{ResNet-50  \\  $\rightarrow$ ResNet-18}}  &\multicolumn{2}{c|}{ \makecell[c]{ResNet-50 \\  $\rightarrow$ ResNet-18} } & \multicolumn{2}{c}{ \makecell[c]{SeNet-154 \\  $\rightarrow$ ResNet-34} } \\ 
\multicolumn{2}{r|}{ Parameter Reduction } &\multicolumn{2}{c|}{None} &\multicolumn{2}{c|}{21.9M $\rightarrow$ 1.7M} &\multicolumn{2}{c|}{63.6M $\rightarrow$ 13.7M} &\multicolumn{2}{c|}{67.6M $\rightarrow$ 14.9M} &\multicolumn{2}{c}{258.4M $\rightarrow$ 38.7M} \\
\hline
\hline
Method &Data & { RMSE} & $\delta_1$  &{ RMSE}  & $\delta_1$  &{ RMSE} &$\delta_1$ &{ RMSE} &$\delta_1$ &{ RMSE} &$\delta_1$ \\
\hline 
Teacher &\multirow{2}{*}{NYU-v2} &{ 0.481} &0.829 &{ 0.481} &0.829 &{ 0.497} &0.824 &{ 0.465}  &0.843 &{ 0.418} &0.878 \\
Student & &{0.481} &0.829 &{ 0.518} &0.802 &{ 0.522} &0.805 & { 0.494} &0.826 &{ 0.459} &0.843\\
\hline
Random noises & \multirow{2}{*}{None}  &{ 1.673} &0.193 &{ 1.702} &0.194 &{ 1.935}  &0.102 &{1.934} &0.112 &{ 1.953} &0.107\\ 
DFAD  & &{ 0.958} &0.402  &{ 1.090} &0.329  &{1.004} &0.382 &{ 1.163} &0.338 &{ 1.208} &0.278\\  
\hline
KD-OOD  &\multirow{2}{*}{SceneNet $\mathcal{X}'_1$}   &{ 0.596} &0.753 &{ 0.648} &0.712  &{ 0.729} &0.660 &{ 0.660}  &0.710 &{0.665} &0.695\\ 
Ours &  &\textbf{{ 0.555}} &\textbf{0.774} &\textbf{{0.600}} &\textbf{0.742} &\textbf{{ 0.676}} &\textbf{0.701} &\textbf{{0.639}} &\textbf{0.722} & \textbf{{ 0.581}} &\textbf{0.759}\\ \hline
KD-OOD  &\multirow{2}{*}{SceneNet $\mathcal{X}'_2$}   &{ 0.590} &0.761 &{ 0.611} &0.742 &{ 0.705} &0.676 &{0.663}  &0.713 &{ 0.605} &0.738\\
    Ours  & & \textbf{{ 0.537}} & \textbf{0.789} &\textbf{{ 0.558}}  & \textbf{0.778}  &\textbf{{ 0.648}} &\textbf{0.726}  &\textbf{{ 0.584}} &\textbf{0.760} & \textbf{{ 0.569}} &\textbf{0.776} \\ \hline
\end{tabular}
\end{center}
\end{table*}

\subsubsection{Datasets}
\paragraph{NYU-v2 \citep{NYUv2}} The NYU-v2 dataset is the benchmark most commonly used for depth estimation. 
 The depth range is 0 to 10 meters.
 It is captured by Microsoft Kinect with an original resolution of $640 \times 480$, and contains 464 indoor scenes. Among them, 249 scenes are chosen for training, and 215 scenes are used for testing.
We use the pre-processed data by Hu et al. \citep{Hu2019RevisitingSI,hu2019analysis}
with approximately 50,000 unique pairs of an image and a depth map with a resolution of 640$\times$480. 
Following most previous studies, we resize the images to 320$\times$240 pixels and then crop their central parts of 304$\times$228 pixels as inputs. 
For testing, we use the official small subset of 654 RGBD pairs.

 \paragraph{ScanNet \citep{dai2017scannet}} ScanNet is a large-scale RGBD dataset that contains 2.5 million RGBD images. The depth range is 0 to 6 meters. We randomly and uniformly select a subset of approximately 50,000 samples from the training splits of 1513 scenes for training and evaluate the models on the test set of another 100 scenes with 17K RGB pairs.
 We apply the same image pre-processing methods, such as image resizing and cropping, as utilized on the NYU-v2 dataset.

\paragraph{SceneNet \citep{mccormac2016scenenet}} SceneNet is a large-scale synthesized dataset that contains 5 Million RGBD indoor images from over 15,000 synthetic trajectories. Each trajectory has 300 rendered frames. The original image resolution is 320$\times$240. Thus, we only apply the center crop to yield an image resolution of 304$\times$228.

We sample two subsets from 1000 indoor scenes of the official validation set.  The two subsets have 50,000 and 300,000 images, respectively, and are denoted by $\mathcal{X}'_1$ and $\mathcal{X}'_2$ in the following texts. 
Detailed information on the datasets used in the experiments is given in Table~\ref{datasets}.

\subsubsection{Models of the Teacher and the Student} 

We choose multiple combinations of the teacher and student models to extensively evaluate our models and methods.
For the first combination, we let the teacher and student models have the same lightweight depth estimation network (LDEN) proposed in \citep{hu2021boosting} built on ResNet-34 \citep{He2016DeepRL} to investigate the performance without model compression. For the second combination,
we use the above ResNet-34 based (LDEN) as the teacher model and the MobileNet-v2 \citep{sandler2018mobilenetv2} based network as the student model in \citep{hu2021boosting}. 
For the next two combinations, the teacher models are implemented using a ResNet-50 \citep{He2016DeepRL} based encoder-decoder network (EDN) \citep{laina2016deeper} and multi-branch feature fusion network (FFN) \citep{Hu2019RevisitingSI}, respectively. Networks of the student models are modified from the teacher networks by replacing ResNet-50 with ResNet-18. For the last combination, the teacher model is a SeNet-154 \citep{hu2018senet} based structure-aware residual pyramid network ((SARPN) \citep{Chen2019structure-aware}. Similarly, the student model is derived from the teacher model by replacing the backbone with a smaller ResNet-34.

To implement the network of the transformation model, we use the dilated convolution \citep{Yu2017} based encoder-decoder network modified from the saliency prediction network \citep{hu2019visualization,hu2019analysis} by adding symmetric skip connections between the encoder and the decoder.

\begin{figure*}[t]
    \centering
    \includegraphics[width = \linewidth]{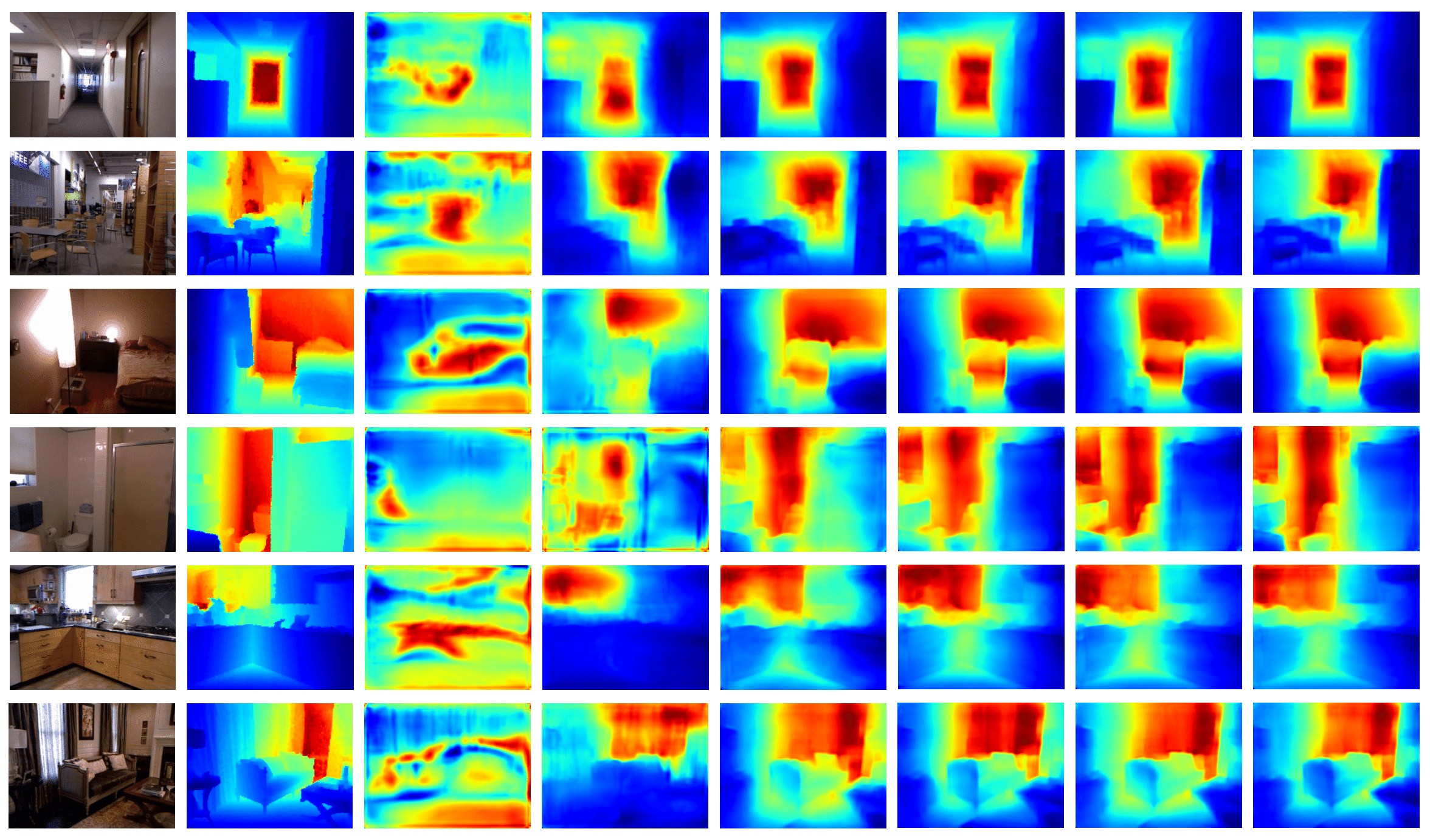} \\
    \begin{tabular}
{p{0.1\textwidth}<{\centering}p{0.1\textwidth}<{\centering}p{0.1\textwidth}<{\centering}p{0.1\textwidth}<{\centering}p{0.1\textwidth}<{\centering}p{0.1\textwidth}<{\centering}p{0.1\textwidth}<{\centering}p{0.1\textwidth}<{\centering}} 

{\footnotesize (a) RGB images.} & {\footnotesize (b) Ground truth.} & {\footnotesize (c) Random noise. } & {\footnotesize (d)  DFAD.} & {\footnotesize (e) KD-OOD with $\mathcal{X}'_1$.} & {\footnotesize (f) Our results with $\mathcal{X}'_1$.} & {\footnotesize (g) KD-OOD with $\mathcal{X}'_2$.} &{\footnotesize (h) Our results with $\mathcal{X}'_2$.} \\
\end{tabular}
\caption{Qualitative comparison of depth maps predicted by different methods on the NYU-v2 test set.}
\label{fig-nyu-pred}
\end{figure*}

\subsubsection{Baselines}
As discussed in Sec.~\ref{premilinary_1}, most of the previous data-free KD methods cannot be applied to depth regression tasks. Thus, we choose DFAD \citep{Fang2019DataFreeAD} as a baseline since this method does not apply the inherent constraint for synthesizing images. Overall, we consider the following methods as baselines for comparison.  
  
\textbf{Teacher:} The teacher model trained on the target dataset.

\textbf{Student:} The student model trained on the target dataset.

\textbf{KD-OOD}: For the sake of comparison, we take KD \citep{Hinton2015DistillingTK} using the OOD simulated data as the strong baseline of our method. It is the first loss term of Eq.~\eqref{eq_learn_S}.

\textbf{Random noise}: The student model is trained via KD with random Gaussian noise. It is also a baseline commonly used for image classification.

\textbf{DFAD:} The student model is trained with data-free adversarial distillation \citep{Fang2019DataFreeAD} that synthesizes images from random noise with adversarial training.

\subsection{Quantitative Comparisons}

\subsubsection{NYU-v2 Dataset}

We first thoroughly evaluate the proposed method on the NYU-v2 dataset. We measure depth maps using the root mean squared error (RMSE) and the thresholding $\delta_1$ accuracy. 
Table~\ref{res_nyu} shows the quantitative results of different methods for various teacher-student combinations where the performance of the student (trained in supervised learning) exhibits an upper bound that we aim to reach.
As seen, distillation with random noise yields the lowest performance, although they are shown to be effective for some toy datasets, \eg, MNIST \citep{LeCun1998GradientbasedLA} and CIFAR-10 \citep{Krizhevsky2009LearningML}, for image classification. 
 Moreover, DFAD has also failed on the task.

Compared to the above methods, KD-OOD demonstrates much better results, showing the advancement of our route that utilizes OOD simulated images. 
In the case of using the smaller set $\mathcal{X}'_1$, it provides $0.165$ mean increase in RMSE (meters) and $0.115$ decrease in $\delta_1$ over the five different model combinations.
 Most importantly, the proposed method outperforms all baselines and attains consistent performance improvement for all different teacher-student combinations.  It yielded $0.115$ and $0.081$ performance degradation in RMSE (meters) and $\delta_1$. Compared to KD-OOD, it achieves $0.05$ meters and $4.0\%$ mean improvement in RMSE and $\delta_1$, respectively.

We then analyze the effect of utilizing the larger set $\mathcal{X}'_2$. As a result, we found a performance boost for both KD-OOD and our method in all experiments when using a larger scale set.
Our method consistently outperforms KD-OOD by 
 $0.06$ meters and $ 4.9\%$ in RMSE and $\delta_1$, respectively.
Besides, our method using $\mathcal{X}'_1$ even outperforms KD-OOD using $\mathcal{X}'_2$.
% , that is to say, we contribute to compressing the data-scale to $1/6$. 

Another observation is that the first two teacher-student model combinations outperform the latter three. The results agree well with previous studies \citep{wang2021knowledge}, which verified that the performance of the student model degrades when the gap in model capacity between them is significant. This problem can be well handled by using an additional assistant model \citep{Mirzadeh2020ImprovedKD}, distilling intermediate features \citep{Liu2020StructuredKD}, multiple teacher models \citep{Tarvainen2017MeanTA}, and the ensemble of distributions \citep{Malinin2020EnsembleDD}. Since it is a common challenge, we leave it as future work.

Figure~\ref{fig-nyu-pred} visualizes a qualitative comparison of different methods. It is seen that random noises produce meaningless predictions, and DFAD estimates coarse depth maps. A closer observation of maps predicted by KD-OOD and our method shows that our proposed method can more accurately estimate depth in local regions. Overall, the quantitative and qualitative results verified the effectiveness of our approach.

\subsubsection{ScanNet Dataset}

To fully evaluate our method, we also test methods using the ScanNet dataset. We use the teacher and student models proposed in \citep{hu2021boosting}. The results are given in Table~\ref{res_scannet}. The final results are highly consistent with those obtained using NYU-v2. Both random noises and DFAD show extremely low accuracy. The proposed method outperforms KD-OOD even using the smaller set. We obtained 10.5$\%$ and 7.5$\%$ improvement in $\delta_1$ and, 0.089 meters and 0.065 meters improvement in RMSE for $\mathcal{X}'_1$ and $\mathcal{X}'_2$, respectively.

\begin{table}[t]
\caption{The results provided by the models on the ScanNet dataset.}
\renewcommand\arraystretch{1.2}
\begin{center}
\label{res_scannet}
\footnotesize
\begin{tabular}
{r|c|cc}
\hline
 Method & Data  &RMSE  & $\delta_1$  \\ \hline
 Teacher Model &\multirow{2}{*}{ScanNet} &0.333 & 0.790\\
 Student Model &  & 0.357 &0.764  \\ \hline
 Random noise & \multirow{2}{*}{None} &1.265 &0.079 \\
 DFAD & &0.725 &0.368 \\ \hline
 KD-OOD  &\multirow{2}{*}{SceneNet $\mathcal{X}'_1$} &0.542 &0.541 \\ 
 Ours & &\textbf{0.453} &\textbf{0.646} \\ \hline
 KD-OOD &\multirow{2}{*}{SceneNet $\mathcal{X}'_2$} &0.477 &0.618 \\   
 Ours & &\textbf{0.412} &\textbf{0.693} \\  \hline
\end{tabular}
\end{center}
\end{table}

\begin{figure*}[t]
    \centering
    \includegraphics[width = \linewidth]{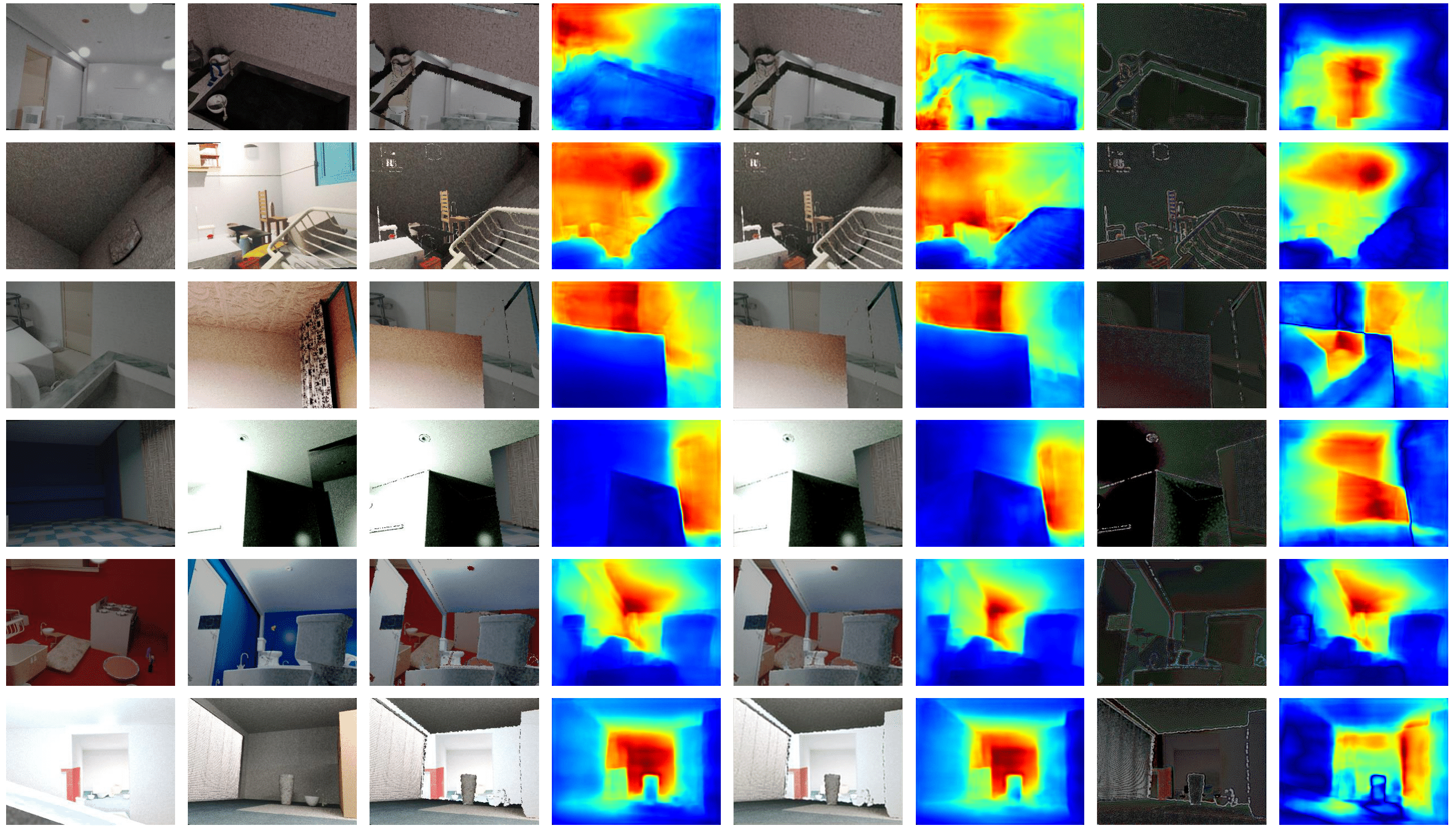}\\
\begin{tabular}
% {cccccccc}
{p{0.1\textwidth}<{\centering}p{0.1\textwidth}<{\centering}p{0.1\textwidth}<{\centering}p{0.1\textwidth}<{\centering}p{0.1\textwidth}<{\centering}p{0.1\textwidth}<{\centering}p{0.1\textwidth}<{\centering}p{0.1\textwidth}<{\centering}} 
    {\small (a) $x'_i$. } & {\small (b) $x'_j$.} & {\small (c) $\hat{x}'$. } & {\small (d) $\mathcal{N}_t(\hat{x}')$. } & {\small (e)  $G(\hat{x}')$. } & {\small (f) $\mathcal{N}_t(G(\hat{x}'))$. } & {\small (g) $d_I$. } &{\small (h) $d_D$. }
\end{tabular}
\caption{Visual comparisons of images and depth maps where (a) and (b) are original images from the simulated set, (c) and (d) are mixed images and estimated depth maps, (e) and (f) denote transformed images of (c) and estimated depth maps, (g) $d_I \triangleq ||\hat{x}'-G(\hat{x}')||_1$ and (h) $d_D \triangleq ||\mathcal{N}_t(\hat{x}')-\mathcal{N}_t(G(\hat{x}'))||_1$ are used to compute image discrepancy and depth discrepancy, respectively.}
\label{fig-nyu-adaption}
\end{figure*}

\subsection{Analyses for the Transformation Model}

Figure~\ref{fig-nyu-adaption} shows some examples of the input and output images of the transformation network as well as their corresponding predictions. In the figure, $x'_i$ and $x'_j$ denote two images randomly selected from the simulated set, and $\hat{x}'$ is the image generated by applying object-wise mixing between $x'_i$ and $x'_j$. $G(\hat{x}')$ denotes the transformed image, \ie, the output of $G$. 
We marked some regions of images in Figure~\ref{fig-nyu-adaption} (c) and (d) with red boxes for better visualization.
By visually comparing $\hat{x}'$ and $G(\hat{x}')$, we observe that $G$ tends to reduce noises and alleviate artifacts around object boundaries such that $G$ can produce more realistic images, mitigating domain gap from the real-world scenarios.  It can be validated by $||\hat{x}'-G(\hat{x}')||_1$ where $|| \cdot ||_1$ is the $\ell_1$ norm (Figure~\ref{fig-nyu-adaption}. (g)) where differences at object boundaries are highlighted. Furthermore, Figure~\ref{fig-nyu-adaption}. (d) and (f) show the predicted depth maps for $\hat{x}'$ and $G(\hat{x}')$, respectively. They demonstrate a clear difference, as observed in Figure~\ref{fig-nyu-adaption}. (h).
We quantify these differences by evaluating the whole set $\mathcal{X}'_1$. As a result, the $\ell_1$-norm of the image and depth difference is 0.156 and 0.227, respectively.

\subsection{Ablation Studies}
We conduct several ablation studies to analyze our approach and provide additional results on the NYU-v2 dataset. Table~\ref{ablation_studies} gives the results. Specifically, we perform several experiments as follows:

\textbf{Without using $\ell_{rec}$:} 
In our original method, we impose the reconstruction consistency between $\hat{x}'$ and $G(\hat{x}')$ to suppress undesirable noises while training the transformation model. 
We relax this constraint and observe that the RMSE and $\delta_1$ dropped to 0.612 and 0.735, respectively.

\textbf{Without using $G$:} We also test the performance while removing the transformation model in the pipeline. We directly perform distillation using $x'$ and mixed images $\hat{x}'$.
As a result, the RMSE and $\delta_1$ dropped to 0.635 and 0.722, respectively.

\textbf{Without using image mixing:} We evaluate the effect without utilizing object-wise image mixing. We feed the images $x'$ to $G$ and apply distillation with both $x'$ and $G(x')$. We find that the RMSE and $\delta_1$ dropped to 0.635 and 0.724, respectively.

\begin{table}[t]
\caption{Results for ablation studies.}
\renewcommand\arraystretch{1.2}
\begin{center}
\label{ablation_studies}
\footnotesize
\begin{tabular}
{l|rr}
% \centering
% {p{0.095\textwidth}<{\centering}p{0.095\textwidth}<{\centering}}
\hline
 & RMSE & $\delta_1$ \\
\hline
Original  &\textbf{0.600} &0.742 \\
Without using $\ell_{rec}$  &0.612 &0.735 \\
Without using $G$  &0.635 &0.722 \\
Without using image mixing &0.635 &0.724 \\
% With $G(x')$ &0.605 &0.741 \\
\hline
\end{tabular}
\end{center}
\end{table}

\section{Conclusion}
\label{conclusion}
We have studied knowledge distillation for monocular depth estimation in data-free scenarios. By first thoroughly analyzing the challenges of the task, we showed that a promising approach to address the challenges is to utilize out-of-distribution (OOD) images as an alternative solution. 
We then empirically and quantitatively verified that i) a high degree of scene similarity, ii) the large-scale size of datasets, and iii) the small magnitude of domain gap contribute to the performance boost of depth distillation methods through detailed experimental analyses with different OOD data. Given the difficulty of data collection in practice, we proposed to utilize simulated images to strengthen the applicability of KD. 
We further presented a novel framework to perform data-free depth distillation with simulated data. 
As a practical solution to the task, we have evaluated the effectiveness of the proposed distillation framework on various depth estimation models and two real-world benchmark datasets. We consider that our framework and results can further inspire future explorations, shedding light on this unexplored problem.

\section*{Acknowledgements}
This work was partly supported by the National Natural Science Foundation of China (62073274, 62106156) and the funding AC01202101103 from the Shenzhen Institute of Artificial Intelligence and Robotics for Society.

\bibliographystyle{cas-model2-names}
\bibliography{egbib}

\end{document}